\documentclass[10pt,twocolumn,letterpaper]{article}

\usepackage{cvpr}              

\usepackage{graphicx}
\graphicspath{{./figures/}}
\usepackage{amsmath}
\usepackage{amssymb}
\usepackage{booktabs}
\usepackage{multirow}
\usepackage{booktabs}
\usepackage{amsfonts}
\usepackage{float}
\usepackage[accsupp]{axessibility}  

\usepackage[pagebackref,breaklinks,colorlinks]{hyperref}

\usepackage[capitalize]{cleveref}
\crefname{section}{Sec.}{Secs.}
\Crefname{section}{Section}{Sections}
\Crefname{table}{Table}{Tables}
\crefname{table}{Tab.}{Tabs.}

\def\cvprPaperID{2677} 
\def\confName{CVPR}
\def\confYear{2023}

\begin{document}
	
	\title{Human Guided Ground-truth Generation for Realistic Image Super-resolution}
	
	\author{Du Chen$^{1}$\thanks{Equal contribution.}, Jie Liang$^{1,2*}$, Xindong Zhang$^{1,2}$, Ming Liu$^{1,3}$, Hui Zeng$^2$ and Lei Zhang$^{1,2}$\thanks{Corresponding author. This work is supported by the Hong Kong RGC RIF grant (R5001-18) and the PolyU-OPPO Joint Innovation Lab.} \\
	$^1$The Hong Kong Polytechnic University, $^2$OPPO Research Institute, $^3$Harbin Institute of Technology\\
	{\tt\small \{csdud.chen, c-ming.liu\}@connet.polyu.hk}, {\tt\small \{liang27jie, cshzeng\}@gmail.com}\\
	{\tt\small \{csxdzhang, cslzhang\}@comp.polyu.edu.hk}
	}
	\maketitle

	\begin{abstract}
		How to generate the ground-truth (GT) image is a critical issue for training realistic image super-resolution (Real-ISR) models. Existing methods mostly take a set of high-resolution (HR) images as GTs and apply various degradations to simulate their low-resolution (LR) counterparts. Though great progress has been achieved, such an LR-HR pair generation scheme has several limitations. First, the perceptual quality of HR images may not be high enough, limiting the quality of Real-ISR outputs. Second, existing schemes do not consider much human perception in GT generation, and the trained models tend to produce over-smoothed results or unpleasant artifacts. With the above considerations, we propose a human guided GT generation scheme. We first elaborately train multiple image enhancement models to improve the perceptual quality of HR images, and enable one LR image having multiple HR counterparts. Human subjects are then involved to annotate the high quality regions among the enhanced HR images as GTs, and label the regions with unpleasant artifacts as negative samples. A human guided GT image dataset with both positive and negative samples is then constructed, and a loss function is proposed to train the Real-ISR models. Experiments show that the Real-ISR models trained on our dataset can produce perceptually more realistic results with less artifacts. Dataset and codes can be found at \url{https://github.com/ChrisDud0257/HGGT}. 
	\end{abstract}
	
	\section{Introduction}
	\label{Introduction}
	
	\begin{figure}[ht]
		\small
		\centering
		\begin{minipage}{0.235\textwidth}
			\includegraphics[width=1\linewidth]{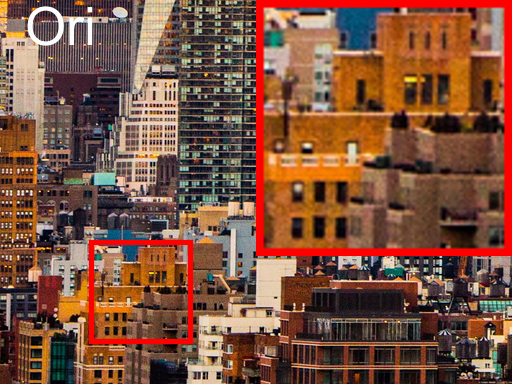}\\
			\includegraphics[width=1\linewidth]{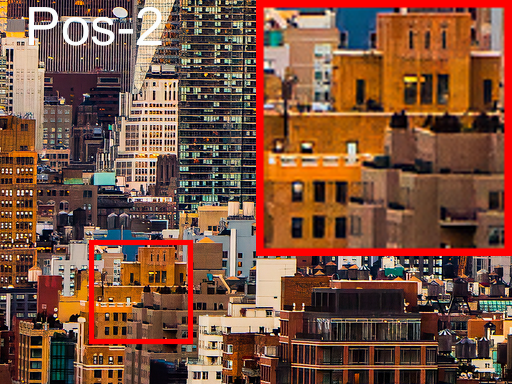}
		\end{minipage}
		\begin{minipage}{0.235\textwidth}
			\includegraphics[width=1\linewidth]{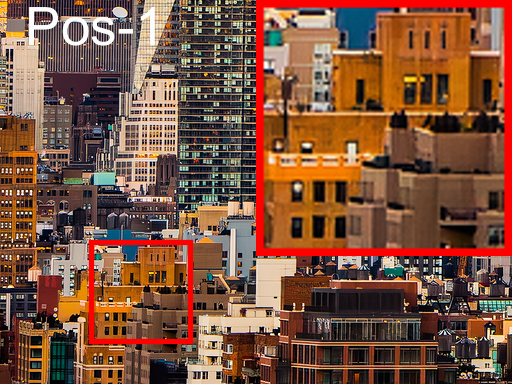}\\
			\includegraphics[width=1\linewidth]{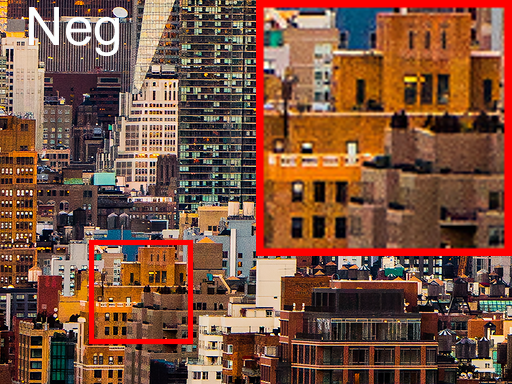}
		\end{minipage}
		\caption{From left to right and top to bottom: one original HR image (Ori) in the DIV2K\cite{agustsson2017ntire} dataset, two of its enhanced positive versions (Pos-1 and Pos-2) and one negative version (Neg). The positive versions generally have clearer details and better perceptual quality, while the negative version has some unpleasant visual artifacts. \textbf{Please zoom in for better observation.}}
		\label{fig:examples of original HR and its enhanced version}
		\vspace{-.6em}
	\end{figure}
	
	Owing to the rapid development of deep learning techniques\cite{he2016deep, johnson2016perceptual, vaswani2017attention, huang2017densely, hu2018squeeze}, the recent years have witnessed the great progress in image super-resolution (ISR)\cite{dong2014learning, dong2015image, dong2016accelerating, kim2016accurate, lai2017deep, lim2017enhanced, ledig2017photo, zhang2018image, zhang2018learning, li2018multi, wang2018esrgan, gu2019blind, bell2019blind, haris2020deep, ma2020structure, zhang2021designing, wang2021real, liang2021swinir, wang2021unsupervised, liang2022details, liang2022efficient, zhang2022efficient}, which aims at generating a high-resolution (HR) version of the low-resolution (LR) input. Most of the ISR models (\eg, CNN\cite{mei2021image, magid2021dynamic} or transformer\cite{chen2021pre,liang2021swinir} based ones) are trained on a large amount of LR-HR image pairs, while the generation of LR-HR image pairs is critical to the real-world performance of ISR models.
	
    Most of the existing ISR methods take the HR images (or after some sharpening operations\cite{wang2021real}) as ground-truths (GTs), and use them to synthesize the LR images to build the LR-HR training pairs. In the early stage, bicubic downsampling is commonly used to synthesize the LR images from their HR counterparts \cite{dong2014learning, dong2015image, kim2016accurate, lim2017enhanced, tai2017image, zhang2018image}. However, the ISR models trained on such HR-LR pairs can hardly generalize to real-world images whose degradation process is much more complex. Therefore, some researchers proposed to collect HR-LR image pairs by using long-short camera focal lengths \cite{chen2019camera, cai2019toward}. While such a degradation process is more reasonable than bicubic downsampling, it only covers a small subspace of possible image degradations. Recently, researchers \cite{gu2019blind, huang2020unfolding, liang2021mutual, zhang2021designing, wang2021real, yue2022blind, luo2022learning, liang2022efficient, zheng2022unfolded} have proposed to shuffle or combine different degradation factors, such as Gaussian/Poisson noise, (an-)isotropic blur kernel, downsampling/upsampling, JPEG compression and so on, to synthesize LR-HR image pairs, largely improving the generalization capability of ISR models to real-world images. 
	
	Though great progress has been achieved, existing  LR-HR training pair generation schemes have several limitations. First, the original HR images are used as the GTs to supervise the ISR model training. However, the perceptual quality of HR images may not be high enough (Fig. \ref{fig:examples of original HR and its enhanced version} shows an example), limiting the performance of the trained ISR models. Second, existing schemes do not consider much human perception in GT generation, and the trained ISR models tend to produce over-smoothed results. When the adversarial losses\cite{ledig2017photo, wang2018esrgan, sajjadi2017enhancenet} are used to improve the ISR details, many unpleasant artifacts can be introduced. 
	
	In order to tackle the aforementioned challenges, we propose a human guided GT data generation strategy to train perceptually more realistic ISR (Real-ISR) models. First, we elaborately train multiple image enhancement models to improve the perceptual quality of HR images. Meanwhile, one LR image can have multiple enhanced HR counterparts instead of only one. Second, to discriminate the visual quality between the original and enhanced images, human subjects are introduced to annotate the regions in enhanced HR images as ``Positive", ``Similar" or ``Negative" samples, which represent better, similar or worse perceptual quality compared with the original HR image. Consequently, a human guided multiple-GT image dataset is constructed, which has both positive and negative samples. 
	
	With the help of human annotation information in our dataset, positive and negative LR-GT training pairs can be generated (examples of the positive and negative GTs can be seen in Fig. \ref{fig:examples of original HR and its enhanced version}), and a new loss function is proposed to train the Real-ISR models. Extensive experiments are conducted to validate the effectiveness and advantages of the proposed GT image generation strategy. With the same backbone, the Real-ISR models trained on our dataset can produce more perceptually realistic details with less artifacts than models trained on the current datasets.
	
	\section{Related Work}
    According to how the LR-HR image pairs are created, the existing ISR methods can be categorized into three major groups: simple degradation based, long-short focal length based, and complex degradation based methods.
	
	\textbf{Simple Degradation based Training Pairs.} Starting from SRCNN\cite{dong2014learning, dong2015image}, most of the deep learning based ISR methods synthesize the LR images from their HR counterparts by bicubic downsampling or direct downsampling after Gaussian smoothing. By using such a simple degradation model to generate a large amount of training data, researchers focus more on  the ISR network module design, such as residual\cite{kim2016accurate}/dense\cite{zhang2018residual} connection, channel-attention\cite{zhang2018image, dai2019second, he2022single}, multiple receptive field\cite{li2018multi, he2019mrfn} or self-attention\cite{chen2021pre, liang2021swinir, zhang2022efficient}.
	The fidelity based measures, such as PSNR and SSIM\cite{wang2004image}, are used to evaluate and compare the performance of different ISR methods. 
	Later on, many works\cite{ledig2017photo, sajjadi2017enhancenet, wang2018recovering, wang2018esrgan, soh2019natural, rad2019srobb, ma2020structure, liang2022details} have been developed to adopt the Generative Adversarial Network (GAN)\cite{goodfellow2020generative} techniques to train Real-ISR models so as to produce photo-realistic textures and details. 
	
	\textbf{Long-short Focal Length based Training Pairs.} Instead of synthesizing LR-HR pairs using simple degradation operators, researchers have also tried to use long-short camera focal length to collect real-world LR-HR pairs. The representative works include CameraSR\cite{chen2019camera} and RealSR\cite{cai2019toward}. The former builds a dataset using DSLR and mobile phone cameras to model degradation between the image resolution and field-of-view. The latter utilizes different focal lengths of the DSLR camera to shot the same scene at different resolutions, and employs an image registration method to crop and align the LR-HR image pairs. Nonetheless, ISR models trained on those datasets might fail when applied to images from different resources (\eg, different degradation, different focal length and cameras).
	
	\begin{figure*}[!ht]
		\centering
		\includegraphics[width=0.98\textwidth]{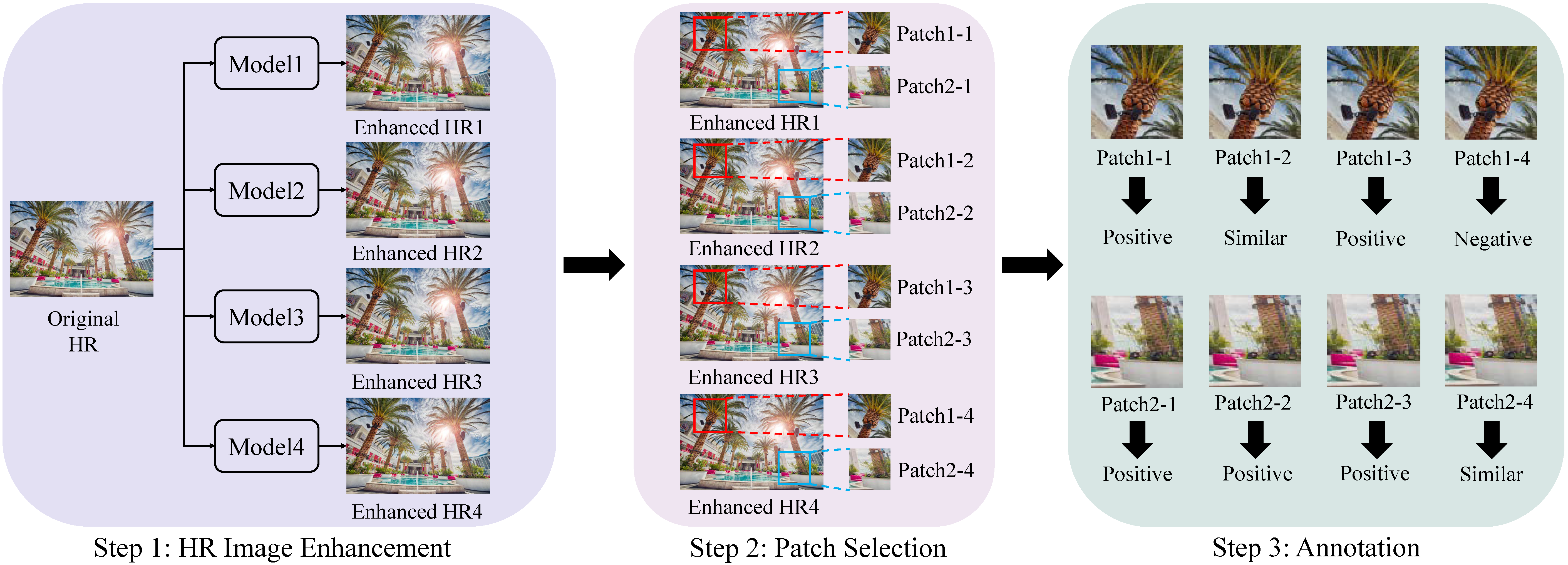}
		\caption{Illustration of our human guided ground-truth (GT) generation process. We first train four image enhancement models to enhance the original high-resolution (HR) image, and then extract the patches which have rich textural and structural details while having certain differences between the original and enhanced versions. Finally, human subjects are involved to annotate the extract patches as ``Positive", ``Similar" and ``Negative" samples.}
		\label{fig:sketch map}
	\end{figure*} 
	
	\textbf{Complex Degradation based Training Pairs.} 
	The image degradation in real-world scenarios can be too complex to model using a simple operator. To enable the Real-ISR models having higher generalization capability, BSRGAN\cite{zhang2021designing} and Real-ESRGAN\cite{wang2021real} have been proposed to synthesize LR-HR training pairs with more complex image degradations. They employ a set of degradation factors, such as different types of noise, blur kernels, scaling factors, JPEG compression, \etc,  to enlarge the degradation space. BSRGAN\cite{zhang2021designing} shuffles and combines different degradations, while Real-ESRGAN\cite{wang2021real} employs a two-stage synthesis progress. In DASR\cite{liang2022efficient}, Liang \etal partitioned the complex degradation space into different levels, and proposed a degradation adaptive method for Real-ISR. 
	
	\textbf{Other Training Pairs.} Beside the above three groups of ISR methods, MCinCGAN\cite{zhang2019multiple} and Pseudo-SR\cite{maeda2020unpaired} utilize unpaired training images to do unsupervised learning. They utilize one or more discriminators to tell the HR GT from the unpaired SR output. AdaTarget\cite{jo2021tackling} employs a transformation CNN block to generate a training-friendly GT from the original GT during the training progress. Nevertheless, the quality of the generated training-friendly GT might not have a good perception quality.
	
	\section{Human Guided Ground-truth Generation}
	
	\subsection{Overview}
	As discussed in Section 2, almost all existing methods \cite{dong2014learning, dong2015image, wang2018esrgan, he2019ode, magid2021dynamic, zhang2021edge, kong2021classsr, luo2022learning} directly take the HR images as the GT to construct the training pairs. Unfortunately, the perceptual quality of many HR images may not be good enough to serve as GTs, limiting the performance trained Real-ISR models. Therefore, we propose to enhance the quality of HR images so that they can better serve as GTs. In particular, human guidance can be introduced in the GT generation process so that perceptually more realistic Real-ISR models can be trained. 
	
	As illustrated in Fig. \ref{fig:sketch map}, the proposed human guided GT generation method has three steps. First, we elaborately train multiple image enhancement models to improve the perceptual quality of HR images. Second, those patches which have enough textural and structural details and have certain differences between the enhanced version and the original version are extracted. Third, human subjects are introduced to discriminate the visual quality between the enhanced patches and the original patch, and label them as ``Positive" (\ie, better quality), ``Similar" (\ie, similar quality) or ``Negative" (\ie, worse quality) samples. In the following subsections, we describe these three steps in detail. 
	
	\subsection{Design of the Enhancement Models}
	\label{Enhancement}
	In order to generate visually more pleasing GTs from the original HR image, we train multiple image enhancement models and apply them to the HR image. To this end, the commonly used DF2K-OST dataset (including 800 high quality images from DIV2K\cite{agustsson2017ntire}, 2650 high quality images from Flickr2K\cite{timofte2017ntire} and 10,324 images from OST\cite{wang2018recovering}) is employed. The original images are denoted by $I^H$, and the low quality ones, denoted by $I^L$, are degraded from $I^H$ by using the following degradation model \cite{zhang2021designing, wang2021real}:
	\begin{equation}
		\label{eq:blind degradation model}
		I^L = [(I^H \otimes K)_R + V]_{J},
	\end{equation}
	where $K$ means isotropic/an-isotropic blur kernel, $R$ means resize operation, $V$ is Gaussian/Poisson noise and $J$ denotes JPEG compression. With ($I^L$, $I^H$) as training pairs, we can train enhancement models. Note that before inputting the low-quality image $I^L$ into the model, we resize it to the size of $I^{H}$ since here we are training enhancement models, where the input and output have the same size.
	
	Considering that the quality of HR image to be further enhanced is generally not bad, we deliberately control the degradation settings in Eq. (1) within weak to middle levels. Otherwise, the learned models can over-enhance the HR images and generate many artifacts. Since the major issues of real world images are noise corruption and blurring, we employ two degradation settings, one focusing on processing slightly high noise and the other focusing on dealing with slightly strong blur. The detailed degradation settings can be found in the \textbf{supplementary file}.
	
	We select one CNN-based network RCAN\cite{zhang2018image} and one transformer-based network ELAN\cite{zhang2022efficient} as the backbones of our enhancer. RCAN\cite{zhang2018image} adopts deep residual learning together with channel-attention\cite{hu2018squeeze}, while ELAN\cite{zhang2022efficient} employs a multi-scale self-attention\cite{vaswani2017attention} block to extract long-range independence. We remove the up-sampling layer in those models since the input and output share the same size in our case. We choose both CNN and transformer as backbones because though transformers have stronger capability in restoring large scale structures and repetitive patterns, CNN can better characterize some small scale and local image details.
	With the two different degradation settings and two different backbones, we train four image enhancement models with $L_1$, perceptual and adversarial losses. The UNet discriminator\cite{wang2021real} is used in adversarial training.
	
	\subsection{Patch Selection and Annotation}
	We apply the trained four enhancement models to 1,600 HR images collected from three representative resources: 1) 800 images from the DIV2K\cite{agustsson2017ntire} dataset; 2) 400 images from Internet which could be used for free, such as Pixabay (\url{https://pixabay.com}) and Unsplash (\url{https://unsplash.com}); 3) 400 images shot by us using mobile phones. Note that though those HR images have high resolution (2K$\sim$4K), they could contain certain noise, blurred details or other real-world degradations, as we shown in Fig. \ref{fig:examples of original HR and its enhanced version}. It is expected that their perceptual quality can be improved by our enhancement models so that they can better serve as GTs in Real-ISR model training. 
	
	After applying the four enhancement models to the 1,600 HR images, we obtain 6,400 enhanced images. However, it is inappropriate to directly take them as GTs. On one hand, many regions in these images are smooth and less informative. On the other hand, there is no guarantee that the enhancement models can always produce perceptually better outputs in all regions. 
	Therefore, we extract patches from those enhanced images and invite human volunteers to label them. In specific, we randomly crop $512*512$ patches from each image with the overlapping area less than $1/2$ of patch area. We then filter out the patches that have large smooth background regions according to the quantity of details and textures, which is measured by the standard deviation (std) of the patch in image domain and the std of high-frequency components in a Laplacian pyramid. At last, we remove the patches on which the difference between the original version and the enhanced version is small (\ie, no much enhancement). The patch selection process avoids the cost of annotating flat patches, and can speed up the training process since flat patches have small gradients. Finally, we select 20,193 groups of patches of $512*512$ size, each group having one original HR patch and 4 enhanced patches.
	
	
	We then invite 60 volunteers with different background to annotate the quality of enhanced patches by comparing them with the original HR patch. A software program, whose interface is shown in the \textbf{supplementary file}, is developed for this purpose. The original patch is positioned at the left side of the screen, while the four enhanced versions are located on the right side in random order. Those patches whose perceptual quality is better than the original one are labeled as ``Positive", and the patches with worse perceptual quality are labeled as ``Negative". In case the quality is tied, the enhanced patch will be labeled as ``Similar". Before annotating, all volunteers are briefly trained to ensure that they will focus on the image perceptual quality (\eg,  sharpness, noise, details, artifacts, \etc) but not on the image content.  
	
	\subsection{Statistics of the Annotated Dataset}
	We invite 60 volunteers to annotate the 20,193 patch groups, each consisting of an original HR patch and 4 enhanced patches. Each group is annotated by 3 different volunteers, and each volunteer is assigned with about 1,010 groups to annotate. In total, we obtain 20,193 groups of annotations, and $20,193\times4\times3=242,316$ annotated patches. The average annotation time for one group is 22.79s.
	
	\textbf{Distribution of the patch annotations.}  
	Tab. \ref{tab:annotation statistics about the overall 20K sets of images} shows the distribution of ``Positive", ``Similar" and ``Negative" labels for each enhancement model, as well as the overall distribution. We see that there are overall 176,042 ``Positive" (72.65\%), 50,904 ``Similar" (21.00\%) and 15,370 ``Negative" (6.35\%) patches. Such statistics imply that our enhancement models improve the visual quality of HR patches in most cases, but there are indeed some bad cases. 
	
	\textbf{Distribution of the final patch labels.} 
	For each of the $20,193\times4=80,772$ enhanced patches, we have three annotations from three different volunteers. We take the majority as the final label of the patch, \ie, if one patch has two or three same annotations, it will be labeled by that annotation. In case the three annotations are different from each other (\ie, one ``Positive", one ``Similar" and one ``Negative"), the final label is marked as ``Similar". Tab. \ref{tab:effective annotation statistics about the overall 20K sets of images} shows the distribution of the final labels of the enhanced patches. We can see that finally there are 63,583 ``Positive" (78.72\%), 14,675 ``Similar" (18.17\%) and 2,514 ``Negative" (3.11\%) patches. Most of the final labels are ``Positive" ones, and only a small portion (3.11\%) are ``Negative" ones. The maximum divergence of ``Positive" labels is 3,329 (5.24\%) between Model 2 and Model 3. The examples of ``Positive", ``Similar" and ``Negative" patches can be found in the \textbf{supplementary file}.
	
	\begin{table}[t]
		\centering
		\caption{The distribution of annotations in our dataset. There are 20,193 groups of patches, while each group consists of an original HR patch and 4 enhanced patches. Each enhanced patch is annotated by 3 different volunteers, resulting in a total of $20,193\times4\times3=242,316$ annotations.}
		\label{tab:annotation statistics about the overall 20K sets of images}
		\resizebox{\linewidth}{!}{
			\begin{tabular}{|c||cccc||c|} 
				\hline
				\multirow{2}{*}{Label} & \multicolumn{4}{c||}{Enhance Model} & \multirow{2}{*}{Total}  \\
				& 1     & 2     & 3     & 4          &                         \\ 
				\hline \hline
				Positive               & 42362 & 39031 & 47251 & 47398      & 176042                  \\
				Similar                & 14623 & 17615 & 10259 & 8407       & 50904                   \\
				Negative               & 3594  & 3933  & 3069  & 4774       & 15370                   \\ 
				\hline \hline
				Total                  & 60579 & 60579 & 60579 & 60579      & 242316                  \\
				\hline
			\end{tabular}
		}
	\end{table}
	
	\begin{table}[t]
		\centering
		\caption{The distribution of final patch labels in our dataset. There are $20,193\times4=80,772$ enhanced patches, each having three annotations. We take the majority annotation label as the final label of each patch.}
		\label{tab:effective annotation statistics about the overall 20K sets of images}
		\resizebox{\linewidth}{!}{
			\begin{tabular}{|c||cccc||c|} 
				\hline
				\multirow{2}{*}{\begin{tabular}[c]{@{}c@{}}Final\\ Label\end{tabular}} & \multicolumn{4}{c||}{Enhance Model} & \multirow{2}{*}{Total}  \\
				& 1     & 2     & 3     & 4          &                         \\ 
				\hline \hline
				Positive                                & 15250 & 13907 & 17236 & 17190     & 63583                   \\
				Similar                                 & 4379  & 5635  & 2517  & 2144      & 14675                   \\
				Negative                                & 564   & 651   & 440   & 859       & 2514                    \\ 
				\hline \hline
				Total                                   & 20193 & 20193 & 20193 & 20193      & 80772                   \\
				\hline
			\end{tabular}
		}
	\end{table}
	
	\begin{table}[h]
		\centering
		\caption{The distribution of the number ($0\sim4$) of final ``Positive" patches per group in our dataset. }
		\label{tab:effective annotation statistics about the sets counts of the overall 20K sets of images}
		\resizebox{\linewidth}{!}{
			\begin{tabular}{|c|ccccc|c|} 
				\hline
				``Positive" Count & 0    & 1    & 2    & 3    & 4     & Total  \\ 
				\hline
				Groups count               & 1267 & 996 & 2616 & 3901 & 11413 & 20193  \\
				\hline
			\end{tabular}
		}
	\end{table}
	
	\textbf{Distribution of the number of final ``Positive" patches per group.} For each group of patches, there can be $0\sim4$ final ``Positive" samples. 
	Tab. \ref{tab:effective annotation statistics about the sets counts of the overall 20K sets of images} shows the distribution of the number of final ``Positive" patches per group. One can see that among the 20,193 groups, 11,413 (56.52\%) groups have 4 ``Positive" patches, 3,901 (19.32\%) have 3 ``Positive" patches, 2,616 (12.95\%) have 2 ``Positive" patches, 996 (4.93\%) have 1 ``Positive" patch, and 1,267 (6.28\%) have none. We will use those ``Positive" patches as ``Positive" GTs, and those ``Negative" patches as ``Negative" GTs to train Real-ISR models. The patches with ``Similar" labels are not employed in our training progress.
	
	\vspace{+3mm}
	\section{Training Strategies}
	\label{training strategies}
	As described in Sec. 3, for an original HR patch, denoted by $I^H$, we may have several (less than 4) positive GTs, denoted by $I^{Pos}$, and several negative GTs, denoted by $I^{Neg}$. To construct the positive or negative LR-GT pairs for Real-ISR model training, we apply the degradation model in Eq. \ref{eq:blind degradation model} to $I^H$ and obtain the corresponding LR image, denoted by $I^L$. (The setting of degradation parameters will be discussed in Sec. \ref{Experimental Settings}). In total, there are 63,583 positive LR-GT pairs $(I^L, I^{Pos})$ and 2,514 negative LR-GT pairs $(I^L, I^{Neg})$. Note that in our dataset, one LR image may correspond to multiple positive GTs or negative GTs. 
	
	\textbf{Training with positive pairs only.}
	\label{Training with All Positive Pairs}
	By removing those groups that do not have any positive GT from the 20,193 training groups, we have 18,926 groups with $1\sim4$ GTs, and 63,583 positive LR-GT pairs to train Real-ISR models. As in previous works \cite{zhang2021designing, wang2021real}, we employ the $L_1$ loss, perceptual loss $L_p$ and GAN loss $L_{GAN}$ to train the model. Since one LR image $I^L$ may have multiple positive GTs, each time we randomly choose one positive GT to calculate the $L_1$, $L_p$ and $L_{GAN}$ losses of the corresponding LR image $I^L$, and update the discriminator and generator networks. The overall training loss is as follows:
	\begin{equation}
		\label{eq:overall loss}
		L_{Total} = \alpha L_1 + \beta L_{p} + \gamma L_{adv},
	\end{equation}
	where $\alpha$, $\beta$ and $\gamma$ are balance parameters.
	
	\textbf{Training with both positive and negative pairs.}
	\label{Training with All Positive and Negative Pairs}
	By filtering out those groups that only contain ``Similar" GTs, we obtain 19,272 groups that have at least one ``Positive" or ``Negative" GT, totally 63,583 positive LR-GT pairs and 2,514 negative LR-GT pairs. When training with the positive GTs, we adopt the same strategy as described above. For each negative LR-GT pair, we introduce a negative loss, denoted by $L_{neg}$, to update the model. 
	
	It is observed that most of the negative GTs have over-sharpened details, strong noise or false details (example images are provided in the \textbf{supplementary file}). 
	Inspired by LDL\cite{liang2022details}, we build a map $\boldsymbol{M}^{Neg}$ to indicate the local residual variation of a negative GT, which is defined as $\boldsymbol{M}^{Neg}_{i,j} =var(\boldsymbol{R}^{Neg}_{i,j}(3,3))^{a}$, where $\boldsymbol{R}^{Neg} = |I^{Neg} - I^H|$ is the residual between the original HR image and the negative GT, $\boldsymbol{R}^{Neg}_{i,j}(3,3)$ is a local $3\times 3$ window of $\boldsymbol{R}^{Neg}$ centered at $(i,j)$, $var$ denotes the variance operation and $a$ is the scaling factor (we set $a$ to $\frac{3}{4}$ in our experiments).
	
	Similarly, we can build a residual variation map $\boldsymbol{M}^{Pos}_{i,j} =var(\boldsymbol{R}^{Pos}_{i,j}(3,3))^{a}$ for the positive GT, where $\boldsymbol{R}^{Pos} = |I^{Pos} - I^H|$. At location $(i,j)$, if the negative residual variation is higher than the positive one, we identify this pixel at $I^{Neg}$ as a truly negative pixel, which should be used to update the model. Therefore, we first define an indication map $\boldsymbol{M}^{Ind}_{i,j}$:
	\begin{equation}
		\label{refine map}
		\boldsymbol{M}^{Ind}_{i,j}=\left\{
		\begin{array}{rcl}
			0, & & {\boldsymbol{M}^{Neg}_{i,j} <= \boldsymbol{M}^{Pos}_{i,j}}\\
			\boldsymbol{M}^{Neg}_{i,j}, & & {\boldsymbol{M}^{Neg}_{i,j} > \boldsymbol{M}^{Pos}_{i,j}}
		\end{array} \right.
	\end{equation}
	and then define the negative loss $L_{neg}$ as follows:
	\begin{equation}
		\label{negative loss}
		L_{neg} = ||\boldsymbol{M}^{Ind}\odot(I^{Neg} - I^{SR})||_{1},
	\end{equation}
	where $\odot$ means dot product. 
 
 Finally, the overall training loss is defined as:
	\begin{equation}
		\label{overall negative loss}
		L_{Total} = \alpha L_1 + \beta L_{p} + \gamma L_{adv} - \delta L_{neg},
	\end{equation}
	where $\alpha$, $\beta$, $\gamma$ and $\delta$ are balance parameters. 
	
	\vspace{+5mm}
	\section{Experimental Results}
	
	\subsection{Experiment Setup}
	\label{Experimental Settings}
	To validate the effectiveness of our human guided GT (HGGT) dataset and the role of negative GTs, we perform two sets of experiments. First, in Sec. 5.2, we train several representative Real-ISR models, such as Real-ESRGAN\cite{wang2021real}, BSRGAN\cite{zhang2021designing}, AdaTarget\cite{jo2021tackling} and LDL\cite{liang2022details} on the DF2K-OST dataset and our HGGT dataset, and compare their performance. Second, in Sec. 5.3, we train two commonly used Real-ISR backbones (RRDB\cite{wang2018esrgan, wang2021real} and SwinIR\cite{liang2021swinir}) on our dataset by using only the postive GTs and using both the positive and negative GTs. 
	
	\textbf{Implementation details.} 
	Before training a model on our dataset, we first pre-train it on the DF2K-OST dataset by using the pixel-wise $\ell_1$ loss to get a stable initialization. Since the original degradation settings in Real-ESRGAN\cite{wang2021real} and BSRGAN\cite{zhang2021designing} is too strong to use in practical ISR applications, we adopt a single-stage degradation process, including blur, noise, down-sampling and JPEG compression with moderate intensities. Detailed settings and visual examples are provided in the \textbf{supplementary file}.
	For the two backbones, RRDB and SwinIR, we utilize the UNet discriminator\cite{wang2021real} for adversarial training, resulting in a RRDB-GAN model and a SwinIR-GAN model. 
	
	We conduct Real-ISR experiments with scaling factor 4 in this paper. We randomly crop training patches of size $256*256$ from the GT images, and resize the corresponding regions in the LR images to $64*64$. The batch size is set to 12 for RRDB backbone and 8 for SwinIR backbone to save GPU memory. We train our model on one NVIDIA RTX 3090 GPU for 300K iterations using the Adam\cite{kingma2014adam} optimizer. The initial learning rate is set to $1e-4$, and we halve it after 200K iterations for RRDB backbone, and 200K, 250K, 275K and 287.5K iterations for SwinIR backbone. The balance parameters $\alpha$, $\beta$, $\gamma$ and $\delta$ in Eq. \ref{overall negative loss} are set to 1, 1, 0.1 and 300, respectively. $\delta$ is set much larger than others because the number of negative GTs is much smaller than positive ones.
	
	\textbf{Testing set.}
	\label{Testing data}
	To evaluate the performance of Real-ISR models trained on our dataset quantitatively, we construct a test set using the same steps as in the construction of our training set. In specific, $100$ patch groups with at least 2 `Positive' GTs are constructed. The input LR patches are generated by using the same degradation process as in the training process. The LR patches together with their GTs are used to quantitatively evaluate the Real-ISR models. We denote this dataset as \textit{Test-100}.
	
	\textbf{Evaluation protocol.}
	For the quantitative evaluation on \textit{Test-100}, we adopt the commonly used PSNR, SSIM~\cite{wang2004image} LPIPS~\cite{zhang2018unreasonable} and DISTS\cite{ding2020image} as quality metrics. Since in \textit{Test-100} one LR image has at least 2 positive GTs, we average the PSNR/SSIM/LPIPS/DISTS scores respectively over the multiple positive GTs as the final scores. For the qualitative evaluation, we invite 12 volunteers to perform subjective assessment, and report the count of preferred Real-ISR models as the user study results.
	
	\begin{figure*}[ht]
		\footnotesize
		\centering
		\begin{minipage}{0.195\textwidth}
			\includegraphics[width=1\linewidth]{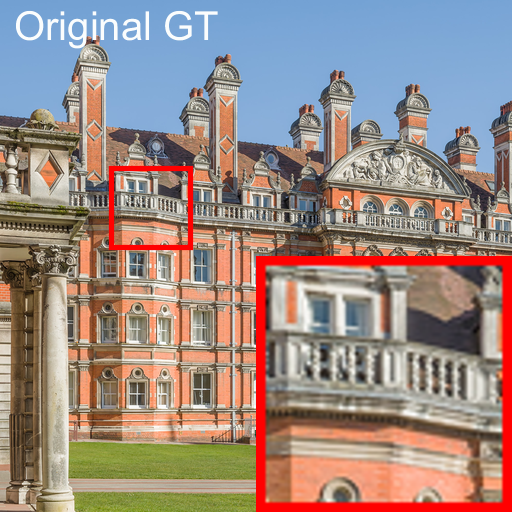}\\
			\includegraphics[width=1\linewidth]{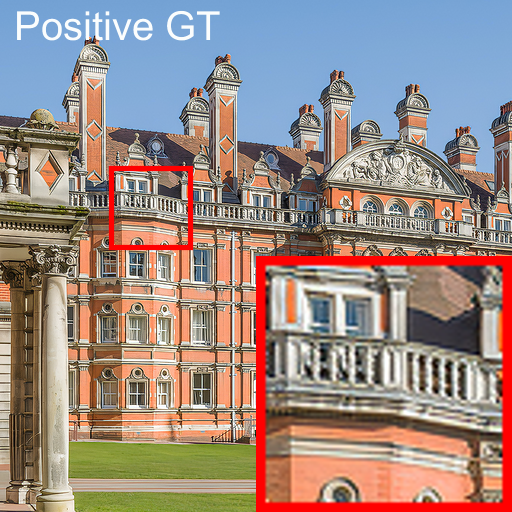}\\
			\includegraphics[width=1\linewidth]{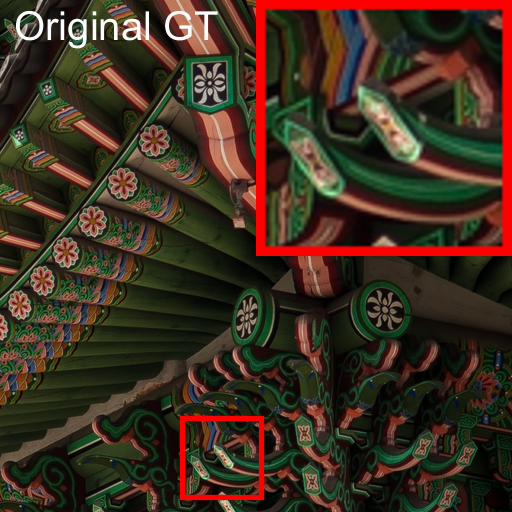}\\
			\includegraphics[width=1\linewidth]{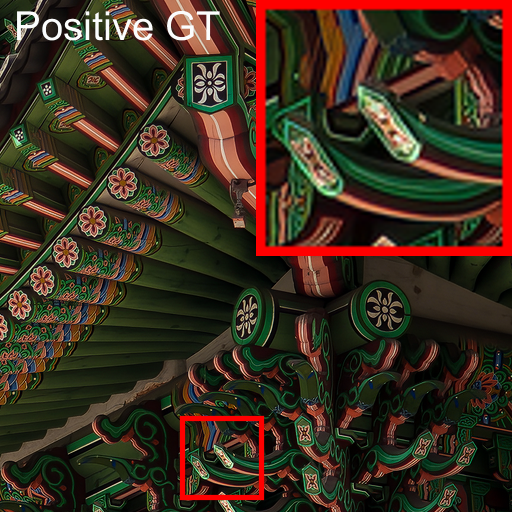}
			\centering{Original\&Positive GT}
		\end{minipage}
		\vspace{0.05cm}
		\begin{minipage}{0.195\textwidth}
			\includegraphics[width=1\linewidth]{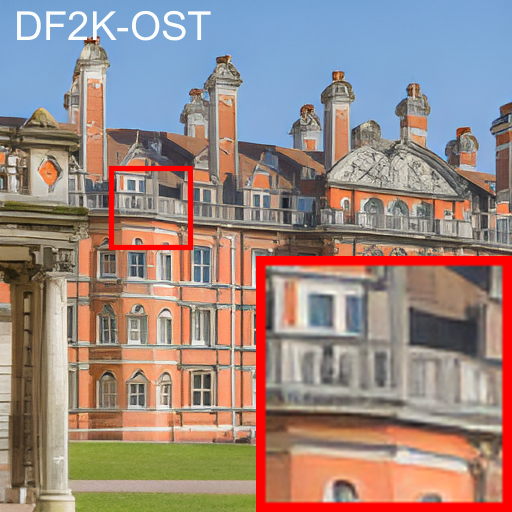}\\
			\includegraphics[width=1\linewidth]{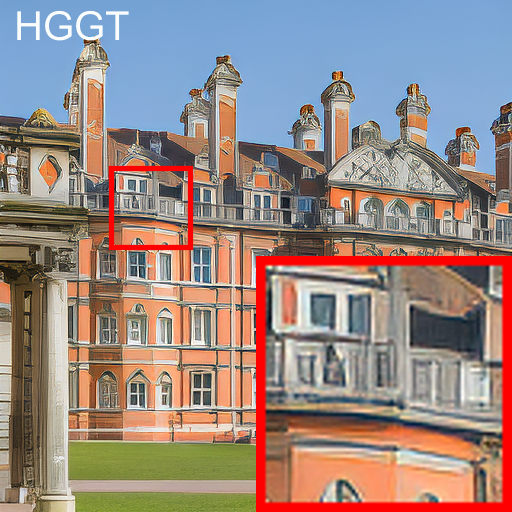}\\
			\includegraphics[width=1\linewidth]{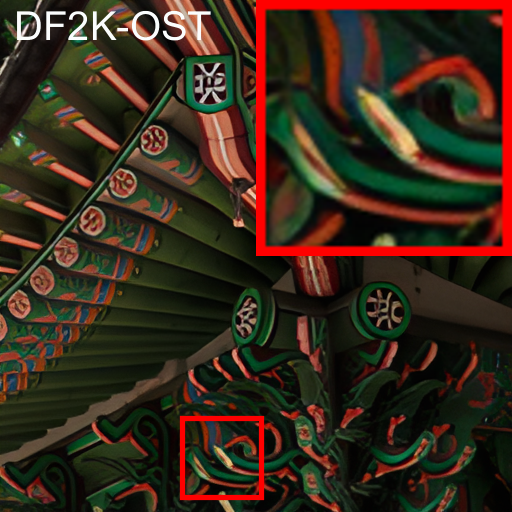}\\
			\includegraphics[width=1\linewidth]{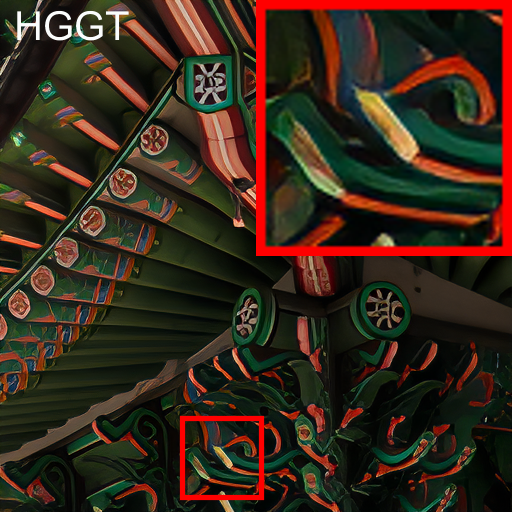}
			\centering{Real-ESRGAN}
		\end{minipage}
		\vspace{0.05cm}
		\begin{minipage}{0.195\textwidth}
			\includegraphics[width=1\linewidth]{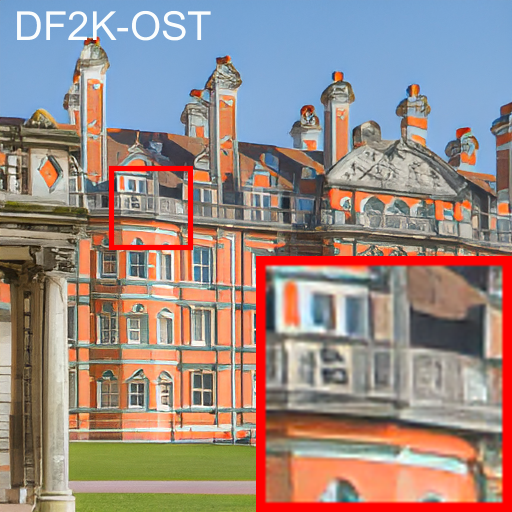}\\
			\includegraphics[width=1\linewidth]{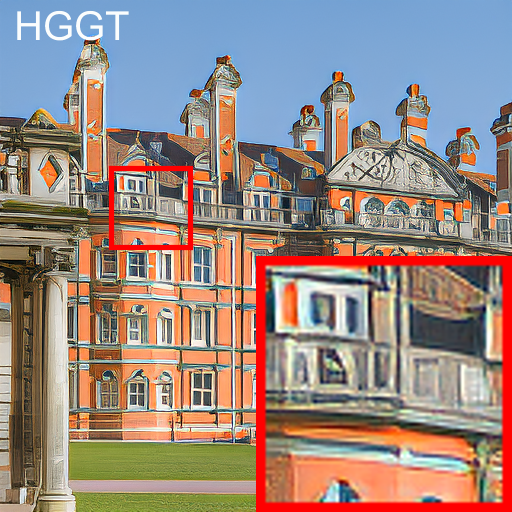}\\
			\includegraphics[width=1\linewidth]{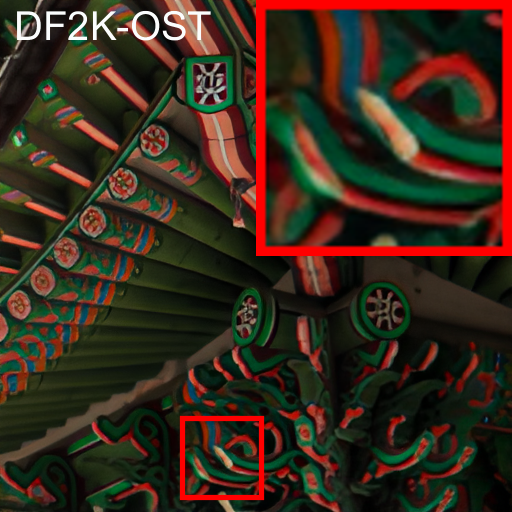}\\
			\includegraphics[width=1\linewidth]{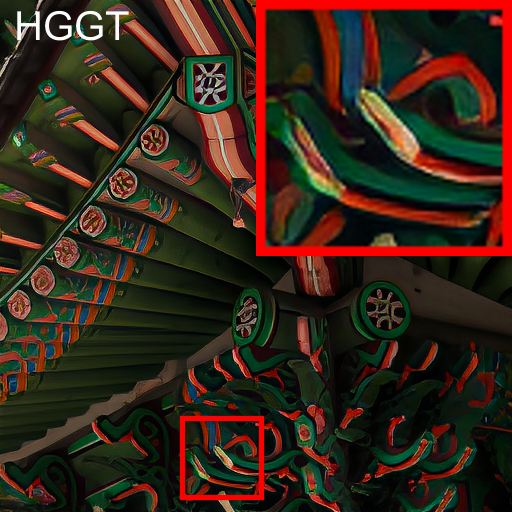}
			\centering{BSRGAN}
		\end{minipage}
		\vspace{0.05cm}
		\begin{minipage}{0.195\textwidth}
			\includegraphics[width=1\linewidth]{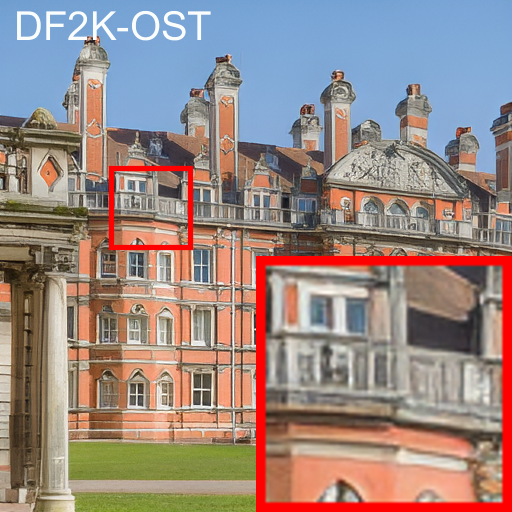}\\
			\includegraphics[width=1\linewidth]{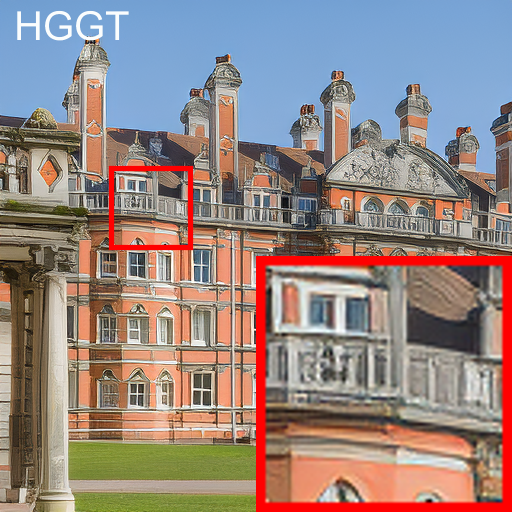}\\
			\includegraphics[width=1\linewidth]{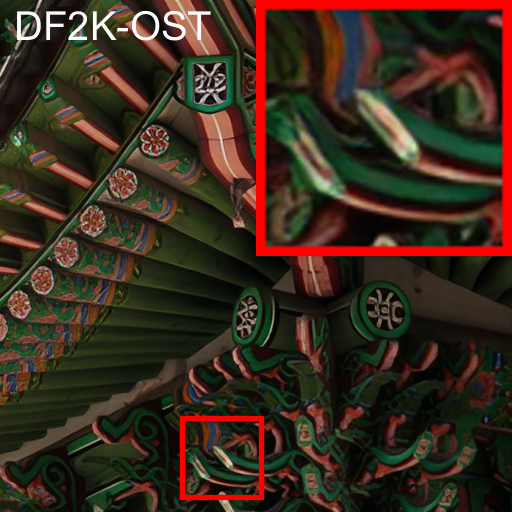}\\
			\includegraphics[width=1\linewidth]{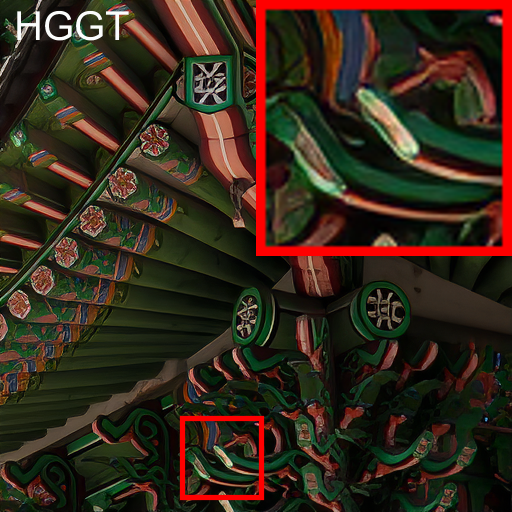}
			\centering{LDL}
		\end{minipage}
		\vspace{0.05cm}
		\begin{minipage}{0.195\textwidth}
			\includegraphics[width=1\linewidth]{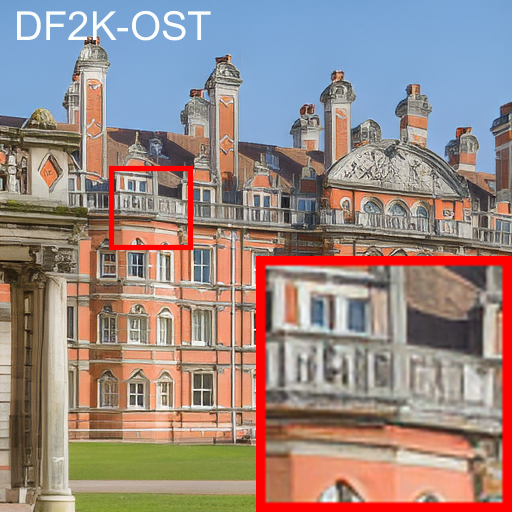}\\
			\includegraphics[width=1\linewidth]{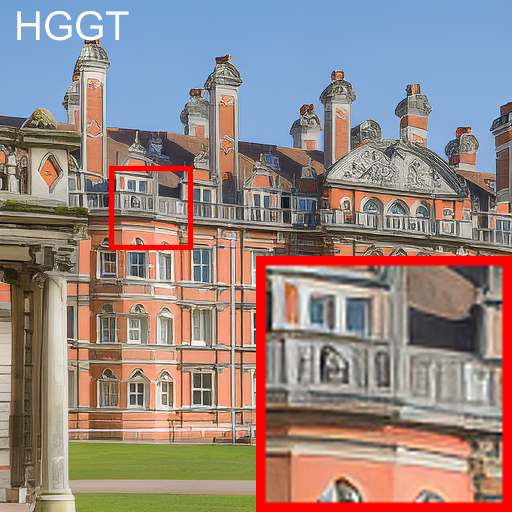}\\
			\includegraphics[width=1\linewidth]{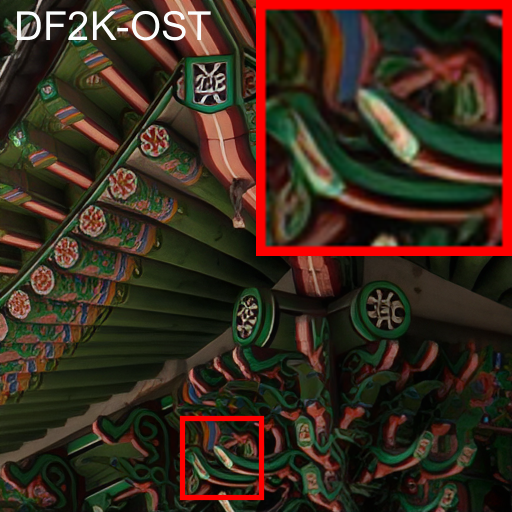}\\
			\includegraphics[width=1\linewidth]{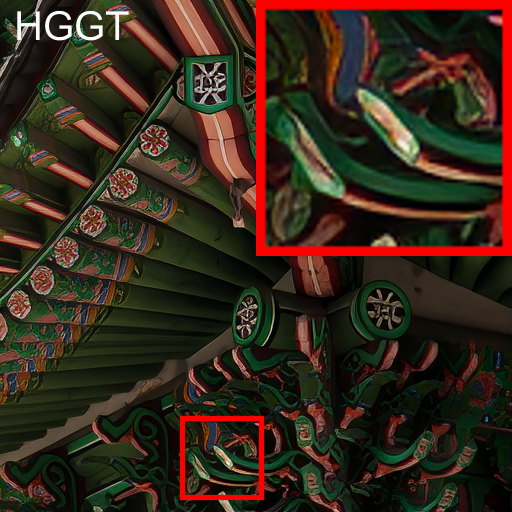}
			\centering{AdaTarget-GAN}
		\end{minipage}
		\vspace{-0.3em}
		\caption{Visual comparison of state-of-the-art models trained on the DF2K-OST and our proposed HGGT datasets. The 1st and 3rd rows show the results of models trained on DF2K-OST, while the 2nd and 4th rows show the results of models trained on ours positive GTs. The left column shows the original GT and the positive GT in our dataset. \textbf{Please zoom in for better observation}.}
		\label{fig:SOTA methods results}
		\vspace{-1.5em}
	\end{figure*}
	
	\begin{figure*}[ht]
		\footnotesize
		\centering
		\begin{minipage}{0.195\textwidth}
			\includegraphics[width=1\linewidth]{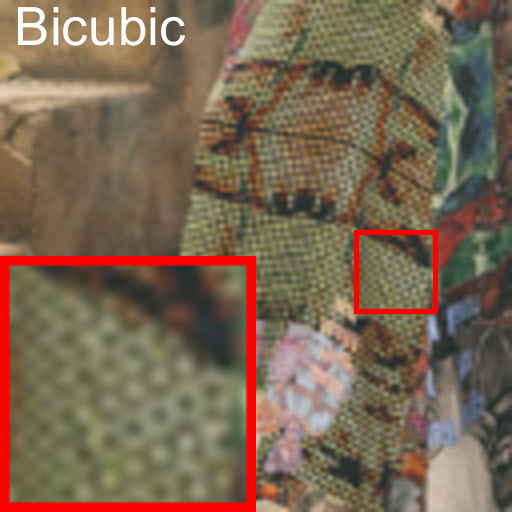}\\
			\includegraphics[width=1\linewidth]{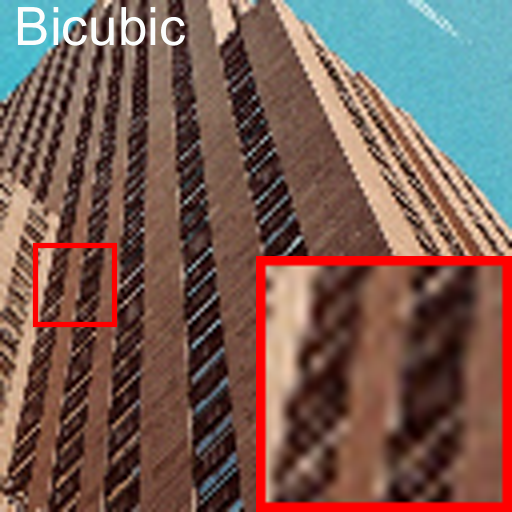}
			\centering{Bicubic}
		\end{minipage}
		\vspace{0.05cm}
		\begin{minipage}{0.195\textwidth}
			\includegraphics[width=1\linewidth]{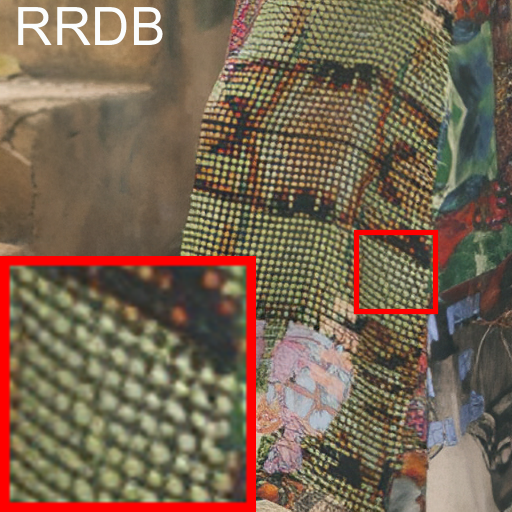}\\
			\includegraphics[width=1\linewidth]{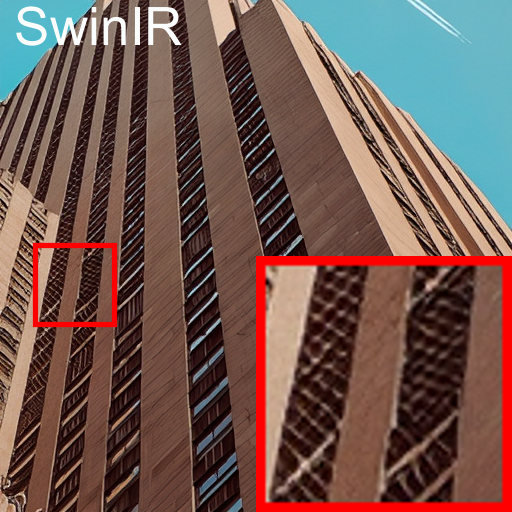}
			\centering{Trained on Ori HR}
		\end{minipage}
		\vspace{0.05cm}
		\begin{minipage}{0.195\textwidth}
			\includegraphics[width=1\linewidth]{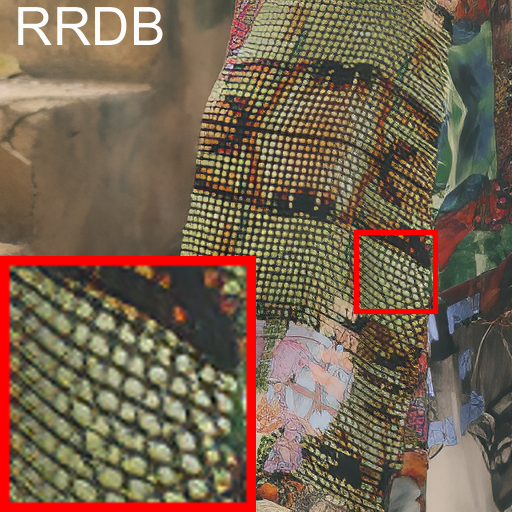}\\
			\includegraphics[width=1\linewidth]{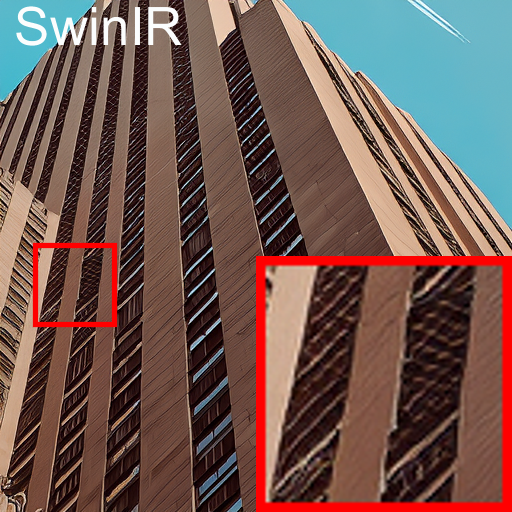}
			\centering{Trained on Pos GT}
		\end{minipage}
		\vspace{0.05cm}
		\begin{minipage}{0.195\textwidth}
			\includegraphics[width=1\linewidth]{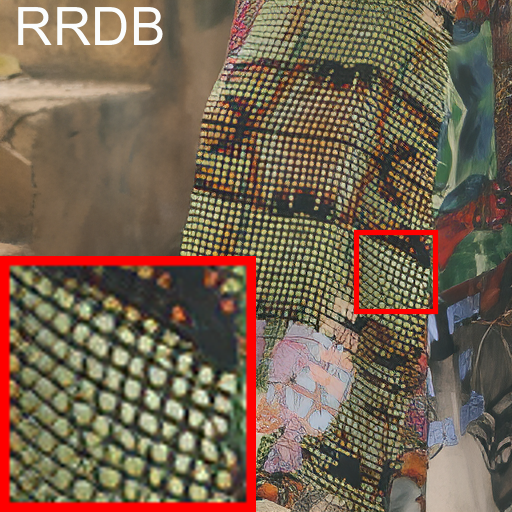}\\
			\includegraphics[width=1\linewidth]{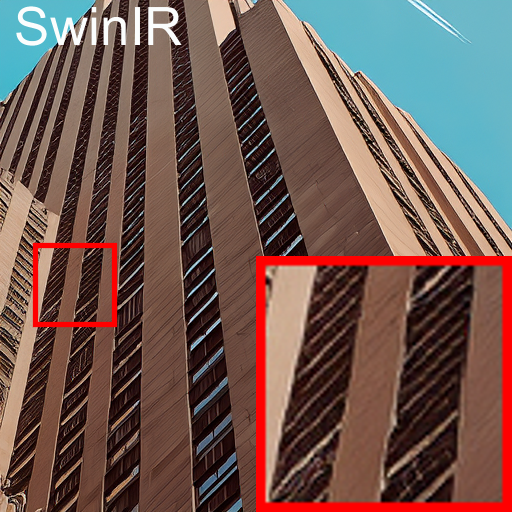}
			\centering{Trained on Pos+Neg GT}
		\end{minipage}
		\vspace{0.05cm}
		\begin{minipage}{0.195\textwidth}
			\includegraphics[width=1\linewidth]{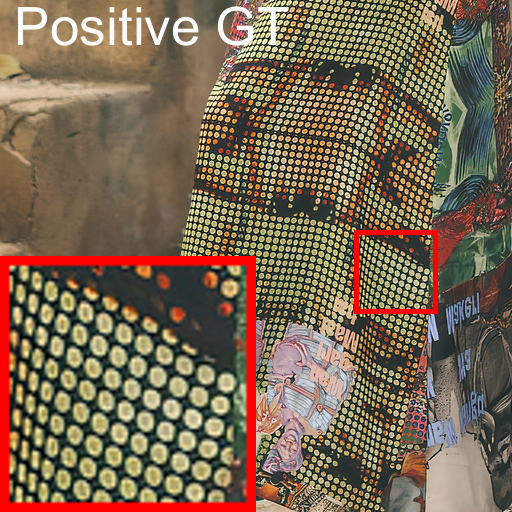}\\
			\includegraphics[width=1\linewidth]{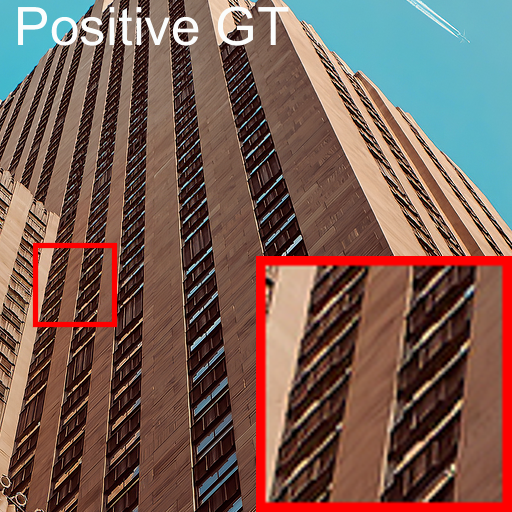}
			\centering{Positive GT}
		\end{minipage}
		\vspace{-.5em}
		\caption{Visualizations of RRDB-GAN ans SwinIR-GAN models trained on the original HR (Ori HR) patches, positive GTs (Pos GT) only, and both positive and negative GTs (Pos+Neg GT) in our HGGT dataset. The top and bottom rows show the results of RRDB-GAN and SwinIR-GAN, respectively. From left to right are the results of bicubic interpolation and the models trained on the Ori HR, Pos GT, Pos+Neg GT, respectively. \textbf{Please zoom in for better observation}.}
		\label{fig:ours methods results}
	\end{figure*}
	
	\subsection{DF2K-OST Dataset vs. Our HGGT Dataset}
	\label{Datasets vesus}
	We first evaluate the effectiveness of the proposed dataset by training representative Real-ISR models respectively on the DF2K-OST dataset and the positive GTs of our HGGT dataset. Four state-of-the-art Real-ISR models are employed, including Real-ESRGAN\cite{wang2021real}, BSRGAN\cite{zhang2021designing}, AdaTarget\cite{jo2021tackling} and LDL\cite{liang2022details}. For Real-ESRGAN and BSRGAN, we adjust the degradation parameters so that the quality of synthesized training LR images is comparable to the LR images in our test set. For AdaTarget and LDL, we use the single-stage degradation as explained in Sec. \ref{Experimental Settings}, and employ the loss functions in the original papers. All models are firstly pre-trained on DF2K-OST with $\ell_1$ loss. The UNet discriminator\cite{wang2021real} is used for adversarial training in our experiments. Quantitative comparison are reported in Table~\ref{tab:quantitive results about SOTA methods} and visual comparisons are shown in Figure~\ref{fig:SOTA methods results}.
	
	As shown in Table~\ref{tab:quantitive results about SOTA methods}, training on our HGGT dataset leads to much better LPIPS/DISTS scores against the DF2K-OST dataset. Specifically, the LPIPS/DISTS scores are significantly improved by $10.14$\%/$12.40$\%, $16.30$\%/$15.27$\%, $17.45$\%/$18.91$\% and $19.23$\%/$21.99$\%, respectively, for Real-ESRGAN, BSRGAN, LDL and AdaTarget-GAN. This indicates a clear advantage of perceptual quality brought by our dataset. Some visual examples are shown in Figure~\ref{fig:SOTA methods results}. One can see that the models trained on our positive GTs can produce perceptually more pleasing results against the models trained on DF2K-OST. The reconstructed images by our dataset have sharper textures and richer details. This is because the original GTs in the DF2K-OST dataset have mixed visual qualities, where a large number of local patches are smooth. 
    In comparison, in our HGGT dataset, the perceptual quality of positive GTs is much enhanced, and the smooth or artifactual patches are mannually removed. These improvements on the training data bring clear advantages to the trained Real-ISR models. More visual results are put in the \textbf{supplementary file}.
	
	As a common problem of GAN-based models, the superior perceptual quality sacrifices the pixel-wise fidelity which is depicted by PSNR and SSIM. This trade-off, which is mainly caused by the ill-posed nature of the image restoration tasks, has been discussed in previous researches\cite{zhang2021designing}. It is well-known that the pixel-wise metrics do not correlate well to the visual quality\cite{ledig2017photo, sajjadi2017enhancenet, soh2019natural}. In addition, in our proposed HGGT dataset, the perceptual quality of GT is improved by using GAN-based enhancement models so that the pixel-wise correlations may not be well-preserved in the data. However, human observers generally prefer the enhanced images in our annotation process, while the perceptually more pleasing results demonstrate the significance of our proposed HGGT dataset in improving the upper bound of the Real-ISR tasks.

\begin{figure}[h]
		\centering
		\includegraphics[width=0.50\textwidth]{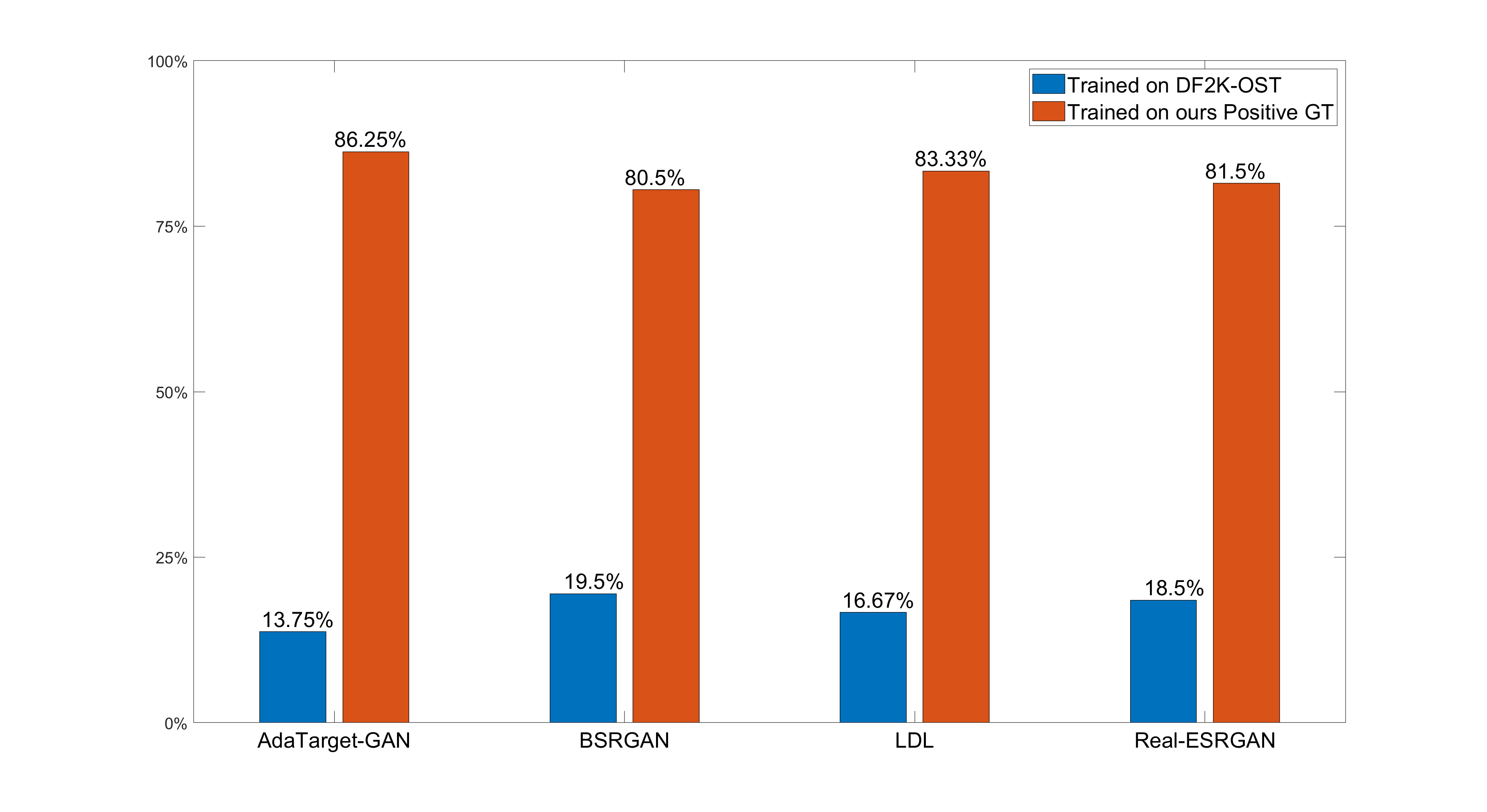}
		\vspace{-2em}
		\caption{User study results on the Real-ISR models trained on the DF2K-OST dataset (the blue bar) and the positive GTs in our HGGT dataset (the red bar).} 
		\label{fig:user study statistic}
	\end{figure}

 The main goal of the proposed HGGT dataset is to improve the perceptual quality of Real-ISR outputs by introducing human perceptions into the training pair generation. We perform a user study to validate the effectiveness of our strategy by inviting 12 volunteers to evaluate the Real-ISR results on the \textit{Test-100} dataset. For each of the four Real-ISR methods, \ie, Real-ESRGAN, BSRGAN, AdaTarget-GAN and  LDL, the two models trained on the DF2K-OST dataset and the positive GTs of our HGGT dataset are compared. Each time, the Real-ISR results of the two models on the same LR input are shown to the volunteers in random order, and the volunteers are asked to chose the perceptually better one based on their evaluation. The statistics of the user study are shown in Fig. \ref{fig:user study statistic}. It should be noted the volunteers invited in this user study do not participate in the annotation process of our dataset.
	
	As shown in Fig. \ref{fig:user study statistic}, the majority of participants (more than 80\% for all tests) prefer the models  trained on our HGGT dataset. This validates the effectiveness of the proposed approach and the dataset, which can be plug-and-play to most of the existing Real-ISR methods and improve their performance by a large margin. For the images where models trained on DF2K-OST are selected, we observe that they mostly contain much flat and smooth regions, and the results of the two models are actually very close.

	\subsection{The Effectiveness of Negative GTs}
	We then evaluate the effectiveness of the negative GTs in our HGGT dataset. We first train the baseline model on the original HR images that are used to build our dataset. Then, we train the models on positive GTs only by using Eq.~\eqref{eq:overall loss}, as illustrated in Section~\ref{training strategies}. Finally, we train the models on both positive and negative GTs by using Eq.~\eqref{overall negative loss}. The CNN-based RRDB and transformer-based SwinIR backbones are used to train Real-ISR models. Due to the limit of space, quantitative comparisons of the trained models are reported in the  \textbf{supplementary file}.

	\begin{table}[h]
	\centering
	\caption{The quantitative results of different Real-ISR models trained on DF2K-OST and our HGGT datasets on \textit{Test-100}.}
	\label{tab:quantitive results about SOTA methods}
	\resizebox{\linewidth}{!}{
		\begin{tabular}{|c||c||c|} 
			\hline
			Method                       & \begin{tabular}[c]{@{}c@{}}Train\\ Dataset\end{tabular} & PSNR/SSIM/LPIPS/DISTS        \\ 
			\hline \hline
			\multirow{2}{*}{Real-ESRGAN} & DF2K-OST                                               & 21.9797/0.6173/0.2593/0.1806  \\
			& Positive GT                                             & 21.5379/0.6078/0.2330/0.1582  \\ 
			\hline \hline
			\multirow{2}{*}{BSRGAN}      & DF2K-OST                                               & 21.7083/0.6092/0.2865/0.1880  \\
			& Positive GT                                             & 20.9037/0.5898/0.2398/0.1593  \\ 
			\hline \hline
			\multirow{2}{*}{LDL}         & DF2K-OST                                               & 22.4724/0.6394/0.2304/0.1676  \\
			& Positive GT                                             & 22.0190/0.6325/0.1902/0.1359  \\ 
			\hline \hline
			\multirow{2}{*}{AdaTarget-GAN}   & DF2K-OST                                               & 22.3944/0.6360/0.2335/0.1687  \\
			& Positive GT                                             & 21.9216/0.6301/0.1886/0.1316  \\
			\hline
		\end{tabular}
	}
	\vspace{-0.5em}
	\end{table}
	
	Visual comparisons are shown in Figure~\ref{fig:ours methods results}, which provides more intuitive evidences on the effectiveness of the annotated negative GTs. As shown in the second column, the models trained on original HR images yield blurry details and irregular patterns, especially on the area with dense textures. This is mainly caused by the low and mixed quality of the original HR images. In contrast, training on our positive GTs can produce much sharper textures and richer details, whereas there remain some false details and visual artifacts (see the windows of the building). Further, training on both positive and negative GTs leads to a more balanced visual performance. Some over-enhanced local pixels can be suppressed, while the textures remain sharp and regular. This is owing to the effective annotation of negative samples, which bring useful human perception guidance into the data for model training. More visual results can be found in the \textbf{supplementary file}.
	

	\section{Conclusion}
	
	In this paper, we elaborately designed a human guided ground-truth (GT) generation method for realistic image super-resolution (Real-ISR). We first trained four image enhancement models to improve the perceptual quality of original high resolution images, and then extracted structural and textural patches from the enhanced images. Finally, human subjects were invited to annotate the perceptual quality of extracted patches as positive and negative GTs, resulting in the human guided ground-truth (HGGT) dataset. The sharper textures and richer details in the positive GTs could largely improve the performance of trained Real-ISR models, while the negative GTs could provide further guidance for the model to avoid generating visual artifacts. Extensive experiments validated the effectiveness of the proposed HGGT dataset and the training strategies.
	
	\textbf{Acknowledgement.} We thank Dr. Lida LI for providing support in GPU server configuration and the many people participating in data collection and annotation.
	
	{\small
		\bibliographystyle{ieee_fullname}
		\bibliography{egbib}
	}
	
\end{document}